# Defeasible Reasoning with Knowledge Graphs[1]

Dave Raggett, email: dsr@w3.org, orcid: 0000-0002-5797-758X

W3C/ERCIM, Sophia Antipolis, France

**Abstract.** Human knowledge is subject to uncertainties, imprecision, incompleteness and inconsistencies. Moreover, the meaning of many everyday terms is dependent on the context. That poses a huge challenge for the Semantic Web. This paper introduces work on an intuitive notation and model for defeasible reasoning with imperfect knowledge, and relates it to previous work on argumentation theory. PKN is to N3 as defeasible reasoning is to deductive logic. Further work is needed on an intuitive syntax for describing reasoning strategies and tactics in declarative terms, drawing upon the AIF ontology for inspiration. The paper closes with observations on symbolic approaches in the era of large language models.

## 1 Defeasible Reasoning

### 1.1 Introduction

The accepted wisdom for knowledge graphs presumes deductive logic as the basis for machine reasoning. In practice, application logic is usually embedded in conventional programming, exploiting scripting APIs and graph query languages, which make it costly to develop and update as application needs evolve.

Declarative approaches to reasoning hold out the promise of increased agility for applications to cope with frequent change. Notation 3 (N3) is a declarative assertion and logic language [1] that extends the RDF data model with formulae, variables, logical implication, functional predicates and a lightweight notation. N3 is based upon traditional logic, which provides mathematical proof for deductive entailments for knowledge that is certain, precise and consistent.

Unfortunately, knowledge is rarely perfect, but is nonetheless amenable to reasoning using guidelines for effective arguments. This paper introduces the Plausible Knowledge Notation (PKN) as an alternative to N3 that is based upon defeasible reasoning as a means to extend knowledge graphs to cover imperfect everyday knowledge that is typically uncertain, imprecise, incomplete and inconsistent.

> *"PKN is to N3 as defeasible reasoning is to deductive logic"*

[1] The work described in this paper was supported by the European Union's Horizon RIA research and innovation programme under grant agreement No. 101092908 (SMARTEDGE)

Defeasible reasoning creates a presumption in favour of the conclusions, which may need to be withdrawn in the light of new information. Reasoning develops arguments in support of, or counter to, some supposition, building upon the facts in the knowledge graph or the conclusions of previous arguments.

As an example, consider the statement: *if it is raining then it is cloudy*. This is generally true, but you can also infer that it is somewhat likely to be raining if it is cloudy. This is plausible based upon your rough knowledge of weather patterns. In place of logical proof, we have multiple lines of argument for and against the premise in question just like in courtrooms and everyday reasoning.

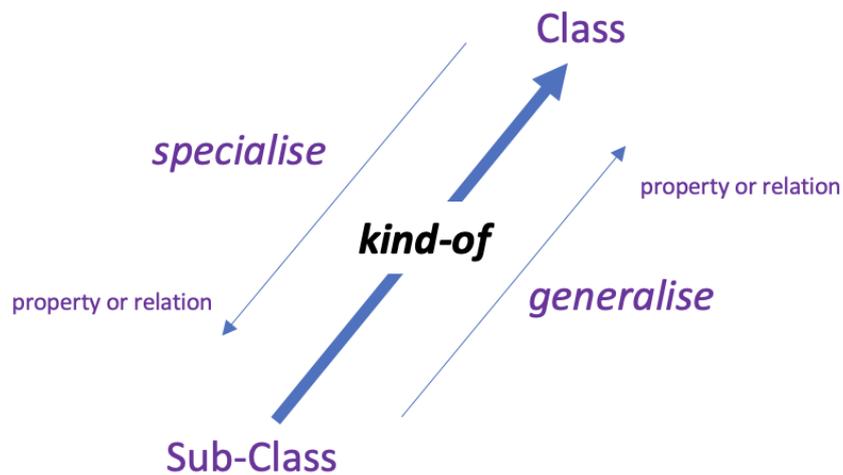

**Fig. 1.** Illustration of how plausible inferences for properties and relations can act as generalizations or specializations of existing knowledge.

The above figure shows how properties and relations involving a class may be likely to apply to a sub-class as a specialization of the parent class. Likewise, properties and relations holding for a sub-class may be likely to apply to the parent class as a generalization. The likelihood of such inferences is influenced by the available metadata. Inferences can also be based on implication rules, and analogies between concepts with matching structural relationships. PKN [2] further supports imprecise concepts:

- fuzzy terms, e.g., cold, warm and hot, which form a scalar range with overlapping meanings.
- fuzzy modifiers, e.g., very old, where such terms are relative to the context they apply to.
- fuzzy quantifiers, e.g., few and many, for queries akin to SPARQL.

PKN represents an evolution from graph databases to cognitive databases, that can more flexibly support reasoning over everyday knowledge. For a web-based demonstrator, see [3].

**1.2  Relation to Previous Work**

The Stanford Encyclopedia of Philosophy entry on argument and argumentation [4] lists five types of arguments: deduction, induction, abduction, analogy and fallacies. Argumentation can be adversarial where one person tries to beat down another, or cooperative where people collaborate on seeking a better joint understanding by exploring arguments for and against a given supposition. The latter may further choose to focus on developing a consensus view, with the risk that argumentation may result in group polarization when people's views become further entrenched.

Studies of argumentation have been made by a long line of philosophers dating back to Ancient Greece, e.g., Carneades and Aristotle. More recently, logicians such as Frege, Hilbert and Russell were primarily interested in mathematical reasoning and argumentation. Stephen Toulmin subsequently criticized the presumption that arguments should be formulated in purely formal deductive terms [5]. Douglas Walton extended tools from formal logic to cover a wider range of arguments [6]. Ulrike Hahn, Mike Oaksford and others applied Bayesian techniques to reasoning and argumentation [7], whilst Alan Collins applied a more intuitive approach to plausible reasoning [8].

Formal approaches to argumentation such as ASPIC+ [9] build arguments from axioms and premises as well as strict and defeasible rules. Strict rules logically entail their conclusions, whilst defeasible rules create a presumption in favor of their conclusions, which may need to be withdrawn in the light of new information.

Arguments in support of, or counter to, some supposition, build upon the facts in the knowledge graph or the conclusions of previous arguments. Preferences between arguments are derived from preferences between rules with additional considerations in respect to consistency. Counter arguments can be classified into three groups. An argument can:

- *undermine* another argument when the conclusions of the former contradict premises of the latter.
- *undercut* another argument by casting doubt on the link between the premises and conclusions of the latter argument.
- *rebut* another argument when their respective conclusions can be shown to be contradictory.

AIF [10] is an ontology intended to serve as the basis for an interlingua between different argumentation formats. It covers information (such as propositions and sentences) and schemes (general patterns of reasoning). The latter can be used to model lines of reasoning as argument graphs that reference information as justification. The ontology provides constraints on valid argument graphs, for example:

*Scheme for Argument from Expert Opinion:*
> *premises*: E asserts that A is true (false), E is an expert in domain D containing A; *conclusion*: A is true (false); *presumptions*: E is a credible expert, A is based on *evidence*; exceptions: E is not reliable, A is not consistent with what other experts assert.

Conflict schemes model how one argument conflicts with another, e.g., if an expert is deemed unreliable, then we cannot rely on that expert's opinions. Preference schemes define preferences between one argument and another, e.g., that expert opinions are preferred over popular opinions. The AIF Core ontology is available in a number of standard ontology formats (RDF/XML, OWL/XML, Manchester OWL Syntax).

PKN defines a simple notation and model for imperfect knowledge. Arguments for and against a supposition are constructed as chains of plausible inferences that are used to generate explanations. PKN draws upon Alan Collins core theory of plausible reasoning [COLLINS] in respect to statement metadata corresponding to intuitions and gut feelings based upon prior experience. This is in contrast to Bayesian techniques that rely on the availability of rich statistics, which are unavailable in many everyday situations.

Recent work on large language models (LLMs), such as GPT-4, have shown impressive capabilities in respect to reasoning and explanations. However, there is a risk of hallucinations, where the system presents convincing yet imaginary results. Symbolic approaches like PKN are expected to play an important and continuing role in supporting semantic interoperability between systems and knowledge graphs. LLMs are trained on very large datasets, and in principle, could be exploited to generate symbolic models in a way that complements traditional approaches to knowledge engineering.

## 2 Plausible Knowledge Notation (PKN)

The Plausible Knowledge Notation is an intuitive lightweight syntax designed to support defeasible reasoning. PKN documents use data types restricted to numbers (as in JSON) and names with optional prefixes.

### 2.1 PKN Statements

PKN supports several kinds of statements: properties, relations, implications and analogies. These optionally include a scope and a set of parameters as metadata. The scope is one or more names that indicate the context in which the statement applies, e.g., ducks are similar to geese in that they are birds with relatively long necks when compared to other bird species. Each parameter consists of a name and a value. Parameters represent prior knowledge as an informal qualitative gut feeling based upon prior experience. Predefined parameters include:

**certainty** - the confidence in the associated statement being true.
**strength** - the confidence in the consequents being true for an implication statement, i.e., the likelihood of the consequents holding if the antecedents hold.

**inverse** - the confidence in the antecedents being true when using an implication statement in reverse, i.e., the likelihood of the antecedents holding if the consequents hold.
**typicality** - the likelihood that a given instance of a class is typical for that class, e.g., that a Robin is a typical song bird.
**similarity** – the extent to which one thing is similar to another, e.g., the extent that they have some of the same properties.
**dominance** - the relative importance of an instance of a class as compared to other instances. For a country, for instance, this could relate to the size of its population or the size of its economy.
**multiplicity** - the number of items in a given range, e.g., how many different kinds of flowers grow in England, remembering that parameters are qualitative not quantitative.

This paper is too short to provide detailed information, so here are a few examples of PKN statements, starting with properties:

```
flowers of Netherlands includes daffodils, tulips (certainty high)
```

Here "flowers" is the descriptor, "Netherlands" is the argument, "includes" is the operator, and "daffodils, tulips" is the referent. In other words, daffodils and tulips are amongst the flowers found in the Netherlands. The metadata indicates that this statement has a high certainty. Next here are two examples of relation statements:

```
Belgium similar-to Netherlands for latitude
Paul close:friend-of John
```

Next here is an implication statement with a locally scoped variable:

```
weather of ?place includes rainy implies weather of ?place includes
cloudy (strength high, inverse low)
```

This example has a single antecedent and a single consequent. Note the use of "?place" as a variable, and metadata for the confidence in using the statement for forward and backward inferences. Next is a couple of examples of analogy statements:

```
leaf:tree::petal:flower
dog:puppy::cat:?
```

Next, here are some examples of queries, which are akin to SPARQL:

```
which ?x where ?x is-a person and age of ?x is very:old
count ?x where age of ?x greater-than 20 from ?x is-a person
few ?x where color of ?x includes yellow from ?x kind-of rose
```

The first query lists the people in the PKN graph who are considered to be very old. The second query counts the number of people older than 20. The third query checks whether there are few yellow roses in the PKN graph.

PKN allows statements to embed sub-graphs for statements about statements, e.g.

```
Mary believes {{John says {John loves Joan}} is-a lie}
```

which models "Mary thinks John is lying when he says he loves Joan."

## 2.2 Fuzzy Knowledge

Plausible reasoning subsumes fuzzy logic as expounded by Lotfi Zadeh in his 1965 paper on fuzzy logic, see [11]. Fuzzy logic includes four parts: fuzzification, fuzzy rules, fuzzy inference and defuzzification.

Fuzzification maps a numerical value, e.g., a temperature reading, into a fuzzy set, where a given temperature could be modelled as 0% cold, 20% warm and 80% hot. This involves transfer functions for each term, and may use a linear ramp or some kind of smooth function for the upper and lower part of the term's range.

Fuzzy rules relate terms from different ranges, e.g., if it is hot, set the fan speed to fast, if it is warm, set the fan speed to slow. The rules can be applied to determine the desired fan speed as a fuzzy set, e.g., 0% stop, 20% slow and 80% fast. Defuzzification maps this back to a numeric value.

Fuzzy logic works with fuzzy sets in a way that mimics Boolean logic in respect to the values associated with the terms in the fuzzy sets. Logical AND is mapped to selecting the minimum value, logical OR is mapped to selecting the maximum value, and logical NOT to one minus the value, assuming values are between zero and one. Plausible reasoning expands on fuzzy logic to support a much broader range of inferences, including context dependent concepts, and the means to express fuzzy modifiers and fuzzy quantifiers.

Here is an example of a scalar range along with the definition of the constituent terms:

```
range of age is infant, child, adult for person
age of infant is 0, 4 for person
age of child is 5, 17 for person
age of adult is 18, age-at-death for person
```

The range property lists the terms used for different categories. The age property for the terms then specifies the numerical range. Additional properties can be used to define the transfer function.

PKN allows terms to be combined with one or more fuzzy modifiers, e.g., "very:old" where very acts like an adjective when applied to a noun. The meaning of modifiers can be expressed using PKN statements for relations and implications, together with scopes for context sensitivity. In respect to old, "very" could either be defined by reference to a term such as "geriatric" as part of a range for "age", or with respect to the numerical value, e.g., greater than 75 years old.

Fuzzy quantifiers have an imprecise meaning, e.g., include few, many and most. Their meaning can be defined in terms of the length of the list of query variable bindings that satisfy the conditions. *few* signifies a small number, *many* signifies a large number, and *most* signifies that the number of bindings for the *where* clause is a large proportion of the number of bindings for the *from* clause.

**2.3  PKN and RDF**

The Resource Description Framework (RDF), see [12], defines a data model for labelled directed graphs, along with exchange formats such as Turtle, query expressions with SPARQL, and schemas with RDF-S, OWL and SHACL. RDF identifiers are either globally scoped (IRIs) or locally scoped (blank nodes). RDF literals include numbers, booleans, dates and strings. String literals can be tagged with a language code or a data type IRI.

The semantics of RDF graphs is based upon Description Logics, see [13] and [14]. RDF assumes that everything that is not known to be true should treated as unknown. This can be contrasted with closed contexts where the absence of some statement implies that it is not true.

Description Logics are based upon deductive proof, whereas, PKN is based upon defeasible reasoning which involves presumptions in favor of plausible inferences, and estimating the degree to which the conclusions hold true. As such, when PKN graphs are translated into RDF, defeasible semantics are implicit and dependent on how the resulting graphs are interpreted. Existing tools such as SPARQL don't support defeasible reasoning.

Consider PKN property statements. The descriptor, argument, operator and referent, along with any statement metadata can be mapped to a set of RDF triples where the subject of the triples is a generated blank node corresponding to the property statement. Comma separated lists for referents and scopes can be mapped to RDF collections.

PKN relations statements can be handled in a similar manner. It might be tempting to translate the relation's subject, relationship and object into a single RDF triple, but this won't work when the PKN relation is constrained to a scope, or is associated with statement metadata. Recent work on RDF 1.2 [15] should help.

PKN implication statements are more complex to handle as they involve a sequence of antecedents and a sequence of consequents, as well as locally scoped variables. One possible approach is to first generate a blank node for the statement, and use it as the subject for RDF collections for the variables, antecedents and consequents.

PKN analogy statements are simpler, although there is a need to be able to distinguish variables from named concepts, e.g. as in "dog:puppy::cat:?".

## 3  Plausible Reasoning and Argumentation

Following the work of Allan Collins, PKN uses qualitative metadata in place of detailed reasoning statistics, which are challenging to obtain. Heuristic algorithms are used to estimate the combined effects of different parameters on the estimated certainty of conclusions. Reasoning generally starts from the supposition in question and seeks evidence, working progressively back to established facts. Sadly, this paper is far too short to go into details and the interested reader should look at the PKN specification.

An open challenge is how to declaratively model strategies and tactics for reasoning rather than needing to hard code them as part of the reasoner's implementation. Further work is needed to clarify the requirements and to evaluate different ways to fulfil those requirements using an intuitively understandable syntax. The AIF ontology would be a useful source of inspiration.

## 4  Summary

This paper introduced PKN as a notation and model for defeasible reasoning with knowledge graphs that include knowledge that is uncertain, imprecise, incomplete and inconsistent. Deductive proof is replaced with plausible arguments for, and against, the supposition in question. This builds on thousands of years of study of effective arguments, and more recently work on argumentation theory. Further work is needed on an intuitive syntax for reasoning strategies and tactics.

Large Language Models have demonstrated impressive capabilities in respect to reasoning and explanations. This raises the question of the role of symbolic approaches such as RDF, N3 and PKN. Deep learning over large corpora has totally eclipsed traditional approaches to knowledge engineering in respect to scope and coverage. However, we are likely to continue to need symbolic approaches as the basis for databases which complement neural networks, just as humans use written records rather than relying on human memory.